\documentclass[review,sort&compress]{elsarticle}

\usepackage{amsmath}
\usepackage{ amssymb }
\usepackage{graphicx}
\usepackage{subcaption}
\usepackage[export]{adjustbox}
\usepackage{color}
\usepackage{algpseudocode}
\usepackage[linesnumbered,ruled,vlined]{algorithm2e}
\usepackage{multirow}

\usepackage{etoolbox}
\makeatletter
\patchcmd{\ps@pprintTitle}
  {Preprint submitted to}
  {Preprint accepted by}
  {}{}
\makeatother

\journal{Applied Soft Computing}

\date{}

\begin{document}

\begin{frontmatter}

\title{Cooperative coevolution of real predator robots and virtual robots in the pursuit domain\tnoteref{mytitlenote}}
\tnotetext[mytitlenote]{This work is partially supported by 
National Key R\&D Program of China under the Grant No. 2017YFC0804002,
National Science Foundation of China under the Grant No. 61761136008, 
Science and Technology Innovation Committee Foundation of Shenzhen under the Grant No. ZDSYS201703031748284, 
Shenzhen Peacock Plan under the Grant No. KQTD2016112514355531, 
Program for Guangdong Introducing Innovative and Entrepreneurial Teams under the Grant No. 2017ZT07X386.}

\author[mymainaddress,mysecondaryaddress]{Lijun Sun}
\ead{11860004@mail.sustech.edu.cn, 13278371@student.uts.edu.au}

\author[mythirdaddress,mymainaddress]{Chao Lyu}
\ead{11849557@mail.sustech.edu.cn}

\author[mymainaddress]{Yuhui Shi\corref{mycorrespondingauthor}}
\cortext[mycorrespondingauthor]{Corresponding author: Yuhui Shi.}
\ead{shiyh@sustech.edu.cn}


\address[mymainaddress]{Shenzhen Key Laboratory of Computational Intelligence, Department of Computer Science and Engineering, Southern University of Science and Technology, China}
\address[mysecondaryaddress]{Centre for Artificial Intelligence, CIBCI Lab, Faculty of Engineering and
Information Technology, University of Technology Sydney, Australia}
\address[mythirdaddress]{Harbin Institute of Technology, China}

\begin{abstract}
The pursuit domain, or predator-prey problem is a standard testbed for the study of coordination techniques. 
In spite that its problem setup is apparently simple, it is challenging for the research of the emerged swarm intelligence. 
This paper presents a particle swarm optimization (PSO) based cooperative coevolutionary algorithm for the (predator) robots, called CCPSO-R, where real and virtual robots coexist in an evolutionary algorithm (EA). 
Virtual robots sample and explore the vicinity of the corresponding real robots and act as their action spaces, while the real robots consist of the real predators who actually pursue the prey robot without fixed behavior rules under the immediate guidance of the fitness function, which is designed in a modular manner with very limited domain knowledge. 
In addition, kinematic limits and collision avoidance considerations are integrated into the update rules of robots.
Experiments are conducted on a scalable swarm of predator robots with 4 types of preys, the results of which show the reliability, generality, and scalability of the proposed CCPSO-R. 
Comparison with a representative dynamic path planning based algorithm Multi-Agent Real-Time Pursuit (MAPS) further shows the effectiveness of CCPSO-R.
Finally, the codes of this paper are public available at: https://github.com/LijunSun90/pursuitCCPSOR.
\end{abstract}

\begin{keyword}
Swarm intelligence, Cooperative coevolution, Particle swarm optimization, Pursuit domain, Virtual robot
\end{keyword}

\end{frontmatter}


\section{Introduction}

The pursuit domain, or predator-prey problem is a classical and interesting research domain which acts as one of the widely used fundamental testbeds for coordination techniques since it was proposed by Benda et al. \cite{benda1986optimal}. 
On one hand, its apparently simple problem setup and flexibility in approaches or concept evaluations lead to both its popularity and the toy domain impression.
On the other hand, it is challenging and thus a good domain for the research of swarm intelligence emerged from the cooperation among robots or agents, which has drawn much attention of researchers on various versions of the pursuit domain. 

At first, greedy coordination strategies were manually designed by Korf \cite{korf1992simple}, part of which were improved by Haynes et al. \cite{haynes1995evolving}.
After that, Haynes et al. \cite{haynes1995evolving,haynes1995strongly,HaynesEvolvingBehavioral,HaynesLearning1998}  improved the pursuit performance using evolutionary algorithms, such as genetic programming (GP) \cite{koza1992genetic}, strongly typed genetic programming (STGP) \cite{montana1995strongly}, and cases learning methods successively.
However, these methods cannot assure $100\%$ capture.
In 2000s, Undeger and Polat \cite{Undeger2010} treated the multi-agent dynamic pursuing problem in partially observable environments with obstacles as a dynamic path planning and task allocation problem and proposed the multi-agent real-time pursuit (MAPS) algorithm. 
Besides, much works have been done in the field of reinforcement learning (RL).
For example, Ishiwaka et al. \cite{ISHIWAKA2003245} investigated the mechanism of the emergence of the predators' cooperative behaviors aiming to capture the prey in the continuous world.
Barrett et al. \cite{Barrett2011Empirical,BARRETT2017132} evaluated the designed single agent in the
ad hoc teamwork and took the pursuit domain as one benchmark task. 
As researches going on, the capture reliability and the efficiency of approaches have both been improved.
A detailed survey on the pursuit domain can be found in \cite{Stone2000}.

In this paper, we deal with the dynamic pursuit domain problem with a scalable swarm of predator robots and types of the prey in bounded diagonal grid worlds.
Different from prior work, this paper treats the pursuit domain as an optimization problem and proposes a particle swarm optimization (PSO) based cooperative coevolutionary (CC) algorithm, called CCPSO-R (R is for robots), where, to the best knowledge of authors, real and virtual robots coexist for the first time in an evolutionary algorithm (EA).
In detail, we have $n$ subpopulations, each of which has the same population size and evolves independently. 
The first individual of each subpopulation always corresponds to a unique real robot, which constitutes the swarm of cooperative real predator robots pursuing the prey.
The rest are virtual robots, which are always deployed around their corresponding real robots, exploring the real robot's vicinity in order to guide the real robot to a more advantageous position under the supervision of the fitness function defined on the pursuit task. 
Hence, in the view of the multi-agent system (MAS), these virtual robots can be seen as the action space for each real robot.
Since the virtual robots only occupy part of a real predator's vicinity, the exploration of virtual robots is actually a sampling rather than an exhausted exploration to a vicinity, which is guided by a proven efficiency swarm intelligence algorithm---PSO \cite{ShiPSO1998}. 
Therefore, the proposed CCPSO-R can be expected to be more efficient and effective.

In addition, the collision avoidance consideration among real robots is integrated into the fitness function design, which not only separates the robotic considerations from the EA itself and is thus different from the robotic PSO (RPSO) \cite{RPSO2011}---the PSO variant specially designed for robots, but also enhances the flexibility of the fitness function by modular design.

Furthermore, unlike previous incremental construction based EA methods and RL algorithms, the proposed CCPSO-R is actually an on-line algorithm which plans one step ahead for each robot and can reliably capture the prey even without the training and learning stage under the immediate guidance of the fitness function.
Meanwhile, similar to the common strategy in RL algorithms, the other real robots (agents in MAS) are treated as parts of the dynamic environment to the current robot without any central commander/controller.

The rest of the paper is organized as follows. 
First, the pursuit domain definition and details adopted here are explained in Section \ref{section_pursuit_domain}. 
Then the proposed CCPSO-R is described in Section \ref{section_algorithm}. 
Experiments, comparisons, corresponding results, and discussions are presented in Section \ref{section_experiments}. 
Finally, conclusions and directions for future researches are given in Section \ref{section_conclusions}.

\section{The pursuit/predator-prey domain}\label{section_pursuit_domain}

Generally speaking, the pursuit domain problem can be considered as a game where predators try to capture the prey with or without coordination. 
However, as summarized in \cite{Stone2000} and mentioned above, the pursuit domain has various versions depending on different combinations of its parameters, such as the type and size of the world, definition of the capture, team size, legal moves and move orders for the predators and prey, distance metric, etc. 
In many researches, a toroidal world is selected to simulate an infinite world, where a robot comes out of one edge will comes in immediately from the opposite edge. 
However, this kind of world is not practical. 
\begin{figure}[h!]
	\hspace*{\fill}
	\begin{subfigure}{0.35\linewidth}
		\includegraphics[width=\linewidth,right]{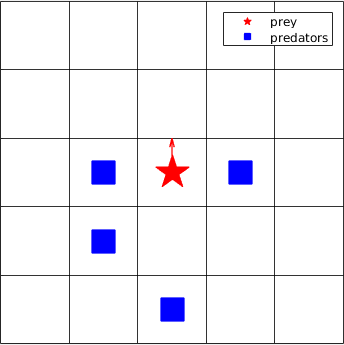}
		\caption{}
  		\label{fig_toroidal_trick_init}
	\end{subfigure}	
	\hfill
	\begin{subfigure}{0.35\linewidth}
		\includegraphics[width=\linewidth,right]{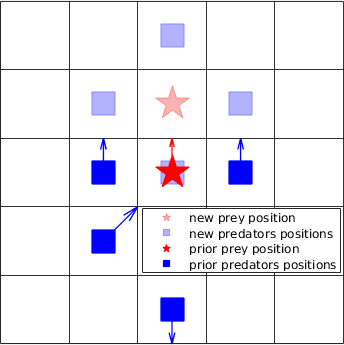}
		\caption{}
  		\label{fig_toroidal_trick_result}
	\end{subfigure}	
	\hspace*{\fill}
	\caption{One trick of the coordination strategy in a toroidal world.}
	\label{fig_toroidal_trick}
\end{figure}
As depicted in Figure \ref{fig_toroidal_trick_init}, if the red pentagram is a linear prey which moves in a straight line towards north and just escapes the nearly encirclement of the predators (blue squares), in the real infinite world, the predators will never catch the prey if they have the same speed. 
But in the toroidal world, if the predators move as shown in Figure \ref{fig_toroidal_trick_result}, they will capture the linear prey in the next step. 
Therefore, in this paper, rather than toroidal worlds, bounded grid worlds are selected, which can at least represent partially, although not all, the real world scenarios, such as an indoor room or an outdoor park with boundaries, etc.

\begin{figure}[h!]
	\centering
    \includegraphics[width=.35\linewidth]{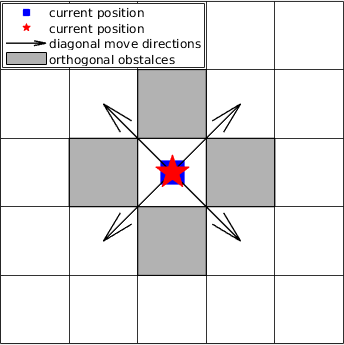}
	\caption{Illustration of the orthogonal obstacles.}
  	\label{fig_orthogonal_obstacles}
\end{figure}

\begin{figure}[h!]
	\centering
	\begin{subfigure}{0.3\linewidth}
		\includegraphics[width=\linewidth]{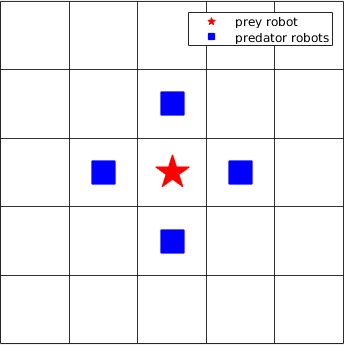}
		\caption{}
  		\label{fig_capture_state_1}
	\end{subfigure}	
	\hfill
	\begin{subfigure}{0.3\linewidth}
		\includegraphics[width=\linewidth]{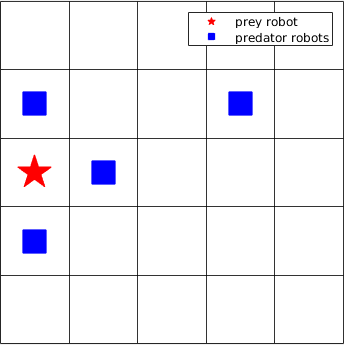}
		\caption{}
  		\label{fig_capture_state_2}
	\end{subfigure}	
	\hfill
	\begin{subfigure}{0.3\linewidth}
		\includegraphics[width=\linewidth]{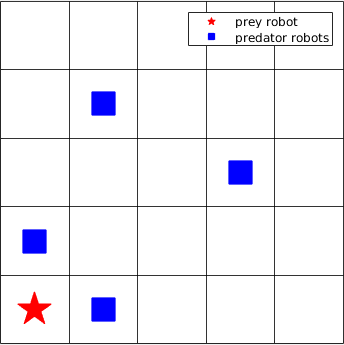}
		\caption{}
  		\label{fig_capture_state_3}
	\end{subfigure}		

	\caption{Illustration of the capture, taking 4 predator robots as an example.}
	\label{fig_capture_state}
\end{figure}

Besides, as classified by Korf \cite{korf1992simple}, the game with a discrete world (grid world here) that only allows horizontal and vertical, totally 4 directions movements, is called the orthogonal game, while the one which allows the horizontal, vertical and diagonal 8 directions move is called the diagonal game. 
Again, towards real applications, the diagonal game is more realistic \cite{korf1992simple} and thus  one of the assumptions of this paper. 
In particular, no collisions are allowed and orthogonal obstacles will be considered when the real (predator or prey) robot moves diagonally, as illustrated in Figure \ref{fig_orthogonal_obstacles}. 
Under these assumptions, the capture can be defined as that every available orthogonal neighbor of the prey robot has been occupied by a predator robot as shown in Figure \ref{fig_capture_state}. 
This may be different from the definitions of some research work, especially the RL algorithms or path planning based methods \cite{Undeger2010}, where the capture is defined as that the position of the prey is occupied by a predator. Thus, the capture is more difficult here. 

As for the other details of the pursuit game, the prey robot always moves first and then the predator robots move one by one in a fixed order, the natural priorities of which can make the collision avoidance control easier and more reliable. 
Besides, the position of any real (prey or predator) robot is visible to all the other robots.
 But there is no explicit communications, i.e. no explicit negotiations or coordination, among the robots. 
In other words, the coordination among real predator robots, or subpopulations, is implicit here. 
In addition, as will be seen later, no fixed behavior rules for the predator robots exist due to the fact that the evolution, or the one step ahead plan, of a predator robot in the dynamic environment is only guided by the fitness function.

\section{Cooperative coevolution of real and virutal robots}\label{section_algorithm}

In this paper, coevolved predators cooperate to encircle a prey, and the evaluation function is called the fitness function in EA \cite{eiben2015introduction}, which is (functionally) identical to an objective function in the optimization filed. 
So, the pursuit domain problem can be treated as an optimization problem in the sense that the goal is to improve the fitness of the pursuit process. 
Concretely, the optimization is conducted by a particle swarm optimization (PSO)  based cooperative coevolutionary (CC) algorithm called CCPSO-R.

\subsection{Fitness function}\label{section_fitness_function}

According to the capture definition in Section \ref{section_pursuit_domain} and the task that a swarm of predator robots needs to encircle a prey robot, the fitness function should subject to the following metrics:
\begin{itemize}
\item \textit{CLOSURE} $f_{closure}$: the prey robot should locate inside the convex hull of the predator robots positions;
\item \textit{SWARM EXPANSE} $f_{expanse}$: the swarm of predator robots should concentrate around the prey robot, i.e., a smaller swarm expanse of the predator robots is preferred;
\item \textit{UNIFORMITY} $f_{uniformity}$: the predator robots should distribute uniformly around the prey robot;
\item \textit{COLLISION AVOIDANCE} $f_{repel}$: collisions among real (predator/prey) robots are not allowed in the practical sense.
\end{itemize}

It is obvious that a single predator robot itself cannot  form a solution. 
In CCPSO-R, a complete solution to the pursuit problem is composed by the positions of all the predator robots. 
However, before formulating the fitness function, a definition needs to be introduced first.

\textbf{Definition of Convex Hull \cite{devadoss2011discrete_and_computational_geometry}:} The \textit{convex hull} of the point set $P$, denoted by conv($P$), is the intersection
of all convex regions that contain $P$. 

\begin{figure}[h!]
	\centering
	\begin{subfigure}{0.3\linewidth}
		\includegraphics[width=\linewidth]{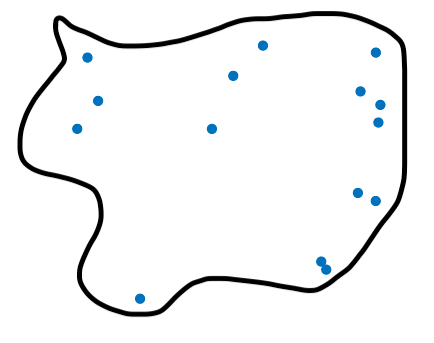}
		\caption{An nonconvex region enclosing $P$.}
  		\label{fig_convex_hull_nonconvex_region}
	\end{subfigure}	
	\hfill
	\begin{subfigure}{0.3\linewidth}
		\includegraphics[width=\linewidth]{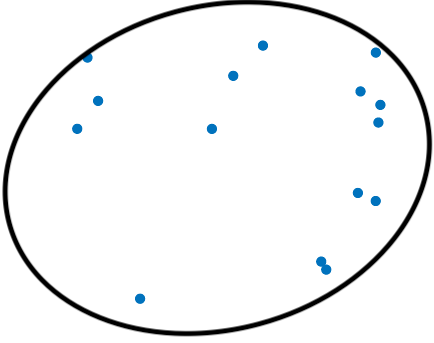}
		\caption{A convex region enclosing $P$.}
  		\label{fig_convex_hull_convex_region}
	\end{subfigure}	
	\hfill
	\begin{subfigure}{0.3\linewidth}
		\includegraphics[width=\linewidth]{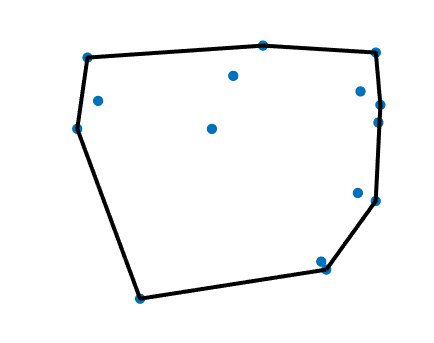}
		\caption{The convex hull of $P$.}
  		\label{fig_convex_hull}
	\end{subfigure}		

	\caption{Illustration of the convex hull of the point set $P$ \cite{devadoss2011discrete_and_computational_geometry}.}
	\label{fig_convex_hull_definition}
\end{figure}

An intuitive illustration of this definition can be found in \cite{devadoss2011discrete_and_computational_geometry} as in Figure \ref{fig_convex_hull_definition}.

Further, we define the function:
\begin{equation}\label{eq_inconv}
inconv(p, conv(P)) \; {\buildrel\rm def\over=} \;
\begin{cases}
0, & \text{if point }p\text{ is in }conv(P) \\
0.5, & \text{if point }p\text{ is on the edge of }conv(P) \\
1, & \text{otherwise}
\end{cases}
\end{equation}

Hence, the fitness function for the $j$th ($j=1,...,N_p$) individual (robot) in the $i$th ($i=1,...,N_s$) subpopulation $p_{robots}^{ij}$ is defined as
\begin{equation}\label{eq_fitness}
f^{ij} = f^{ij}_{repel} \cdot (f^{ij}_{closure} + f^{ij}_{expanse} + f^{ij}_{uniformity})
\end{equation}
where
\begin{equation}\label{eq_fit_repel}
f^{ij}_{repel} = 
\begin{cases}
e^{-2 \cdot (NND^{ij} - D_{min})}, & \text{if }NND^{ij} < D_{min} \\
1, & \text{else}
\end{cases}
\end{equation}
corresponds to the above \textit{COLLISION AVOIDANCE} metric, 
\begin{equation}\label{eq_fit_closure}
f^{ij}_{closure} = inconv(p_{prey}, conv({p_{robots}^{11},...,p_{robots}^{ij},...,p_{robots}^{N_{s}1}}))
\end{equation}
corresponds to the above \textit{CLOSURE} metric, 
\begin{equation}\label{eq_fit_expanse}
f^{ij}_{expanse} = \frac{1}{N_s}(\sum_{k=1,k \ne i}^{N_s}{|p_{robots}^{k1} - p_{prey}|} + |p_{robots}^{ij} - p_{prey}|)
\end{equation}
corresponds to the above \textit{SWARM EXPANSE} metric, 
and
\begin{equation}\label{eq_fit_uniformity}
f^{ij}_{uniformity} = std
\left(\left[
\begin{array}{cc}
	N_{11} & N_{12}  \\
	N_{21} & N_{22}
\end{array}
\right]\right)
\end{equation}
corresponds to the above \textit{UNIFORMITY} metric.

\begin{figure}[htbp]
	\hspace*{\fill}
	\begin{subfigure}{.35\linewidth}
		\includegraphics[width=\linewidth]{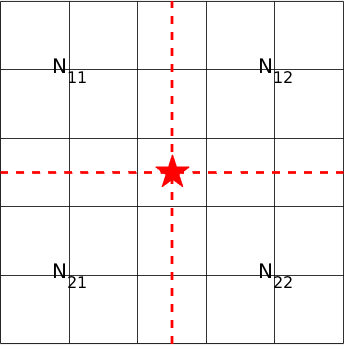}
  		\caption{4 bins split around the prey robot.}
  		\label{fig_uniformity_split}
	\end{subfigure} 
	\hfill
	\begin{subfigure}{.35\linewidth}
		\includegraphics[width=\linewidth]{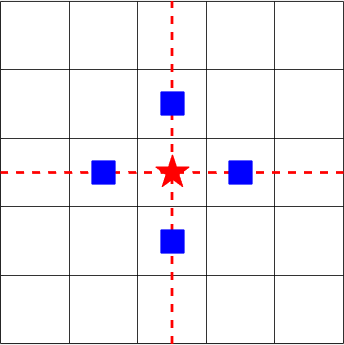}
  		\caption{An example for the uniformity assessment.}
  		\label{fig_uniformity_assessment_example}
	\end{subfigure}	
	\hspace*{\fill}
	\caption{Illustration of the uniformity assessment.}
	\label{fig_uniformity_assessment}
\end{figure}

In the formulas, 
$NND^{ij}$ is the nearest neighbor distance, i.e., the minimum of the pairwise Euclidean distances between the $j$th individual in the $i$th subpopulation and all the real predator robots in the other subpopulations; 
$D_{min}$ is a specified secure distance for collision avoidance; 
$p_{prey}$ is the position of the prey robot; 
$p_{robots}^{ij}$ is the position of the $j$th robot in the $i$th subpopulation; 
$std(\cdot)$ stands for the standard deviation function; 
and $N_{kh} (k=1,2; h=1,2)$ is the counts of the real predator robots in the $(k,h)$-th bin out of the overall 4 bins split by the horizontal and vertical lines which intersect at the position of the prey robot, as shown in Figure \ref{fig_uniformity_assessment}. 
Note that, the number of the real predator robots on the split lines is divided by 2 and equally assigned to the two adjacent bins. Hence, $N_{11}=N_{12}=N_{21}=N_{22} = 1$ for the example in Figure \ref{fig_uniformity_assessment_example}.
\begin{figure}[htbp]
	\hspace*{\fill}
	\begin{subfigure}{.35\linewidth}
		\includegraphics[width=\linewidth]{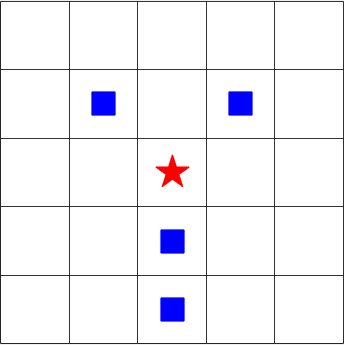}
  		\caption{The uniformity assessment is 0 by equation (\ref{eq_fit_uniformity}) based on the split of Figure \ref{fig_uniformity_split}.}
  		\label{fig_uniformity_deadlock}
	\end{subfigure} 
	\hfill
	\begin{subfigure}{.35\linewidth}
		\includegraphics[width=\linewidth]{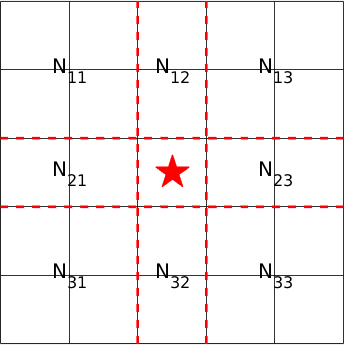}
  		\caption{An alternative split method for the uniformity assessment of equation (\ref{eq_fit_uniformity_alternative}).}
  		\label{fig_uniformity_alternative_split}
	\end{subfigure}	
	\hspace*{\fill}
	\caption{Illustration of the alternative uniformity assessment.}
	\label{fig_uniformity_alternative}
\end{figure}
However, the formula (\ref{eq_fit_uniformity}) cannot always give the objective uniformity assessment that is consistent with a human's subjective judgment, as the deadlock phenomenon shown in Figure \ref{fig_uniformity_deadlock}.
In this scenario, the prey robot always keeps still in the center of the map, while the predator robots start to encircle the prey from randomly generated initial positions and stop forever since the game state shown in Figure \ref{fig_uniformity_deadlock}, which is obviously not the expected capture state.
So, the deadlock phenomenon is a game state where the pursuit task hasn't been accomplished but all the predator robots stop forever as if they are locked.
One reason of the deadlock phenomenon is the fitness function wrongly evaluates an intermediate game state as fittest, i.e., the task is finished.
Therefore, to design a better fitness function, in the uniformity assessment formula, an alternative space split strategy is performed as shown in Figure \ref{fig_uniformity_alternative_split}, and the following uniformity assessment will replace equation (\ref{eq_fit_uniformity}) in such situations:
\begin{equation}\label{eq_fit_uniformity_alternative}
f^{ij}_{uniformity} \\
= std([N_{12},N_{21},N_{23},N_{32}]) + std([N_{11},N_{13},N_{31},N_{33}]),
\end{equation}
the first and second part of which are the axial and diagonal uniformity assessments, respectively.

\begin{figure}[htbp]
	\centering
	\includegraphics[width=\linewidth]{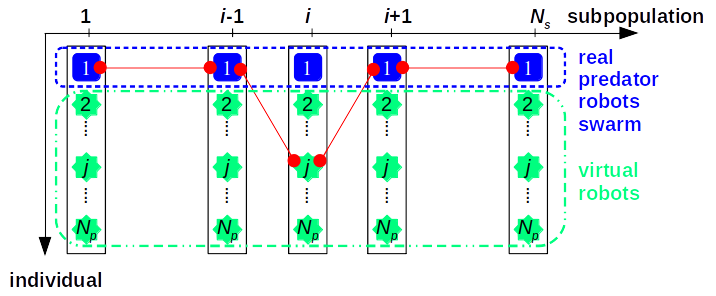}
	\caption{Illustration of the fitness evaluation for $p_{robots}^{ij}$.}
	\label{fig_fitness_evaluation_ij}
\end{figure}

To be clearer, the fitness evaluation of $p_{robots}^{ij}$, i.e., the $j$th ($j=1,...,N_p$) individual (robot) in the $i$th ($i=1,...,N_s$) subpopulation, is illustrated in Figure \ref{fig_fitness_evaluation_ij}.

\subsection{The proposed CCPSO-R algorithm}

In CCPSO-R, there are $N_s$ independently evolved subpopulations with subpopulation size $N_p$, and the first individual of each subpopulation represents a unique real  robot while the others represent virtual robots. 
All the real robots consist of the predator robots swarm which actually pursue the prey robot in the grid world, while the virtual robots are to explore the vicinity of the corresponding real predator robot in its subpopulation and guide the predator robot to a better position.
So in this sense, virtual robots can be seen as the action space of the corresponding real predator robot.
The real predator robot chooses its locally optimal action, but in terms of the global benefit of the whole swarm of predators. 
That is, the evaluation of a robot position is conducted by considering the rest real predator robots positions in the other subpopulations.
Since the proposed algorithm works in the modes of cooperative coevolutionary algorithms (CCEAs), it is called the cooperative coevolutionary PSO for robots (CCPSO-R), as illustrated in Algorithm \ref{algorithm_CCPSO-R}, which will be explained in detail from 3 aspects: the update rules, the fitness evaluation, and the diversity maintenance mechanism in the following.

\begin{algorithm}[h!]
\caption{CCPSO-R}\label{algorithm_CCPSO-R}
Initialization\\
\While{the prey is not captured and time limit is not reached}{
	\For{each subpopulation}{
		Re-evaluate the subpopulation due to environmental changes\\
		\For{each virtual robot}{
			Update its velocity and position using (\ref{eq_velocity_virtual}) and (\ref{eq_position_virtual})\\
			Evaluate the fitness together with the rest real predator robots 
		}
		\If{unique virtual robots $< T_v$}{
			Re-initiate and re-evaluate the virtual robots
		}
		Update the velocity and position of the real predator robot using (\ref{eq_velocity_real}) and (\ref{eq_position_real})\\
		Evaluate the fitness of the real predator robot together with the rest real predator robots\\
		\uIf{the real predator robot becomes the global best}{
			Re-initiate and re-evaluate the virtual robots
		}
		\ElseIf{the predator robot gets trapped in a deadlock}{
			Add a random noise to the real predator robot position\\
			Re-evaluate the whole population
		}	
	}
	\If{the real predator robots swarm get trapped in a local optimum}{
		Add random noises to all the real predator robots positions\\
		Re-evaluate the whole population
	}
}
\end{algorithm}

\subsubsection{Update rules}

\begin{figure}[h!]
	\centering
	\includegraphics[width=.45\linewidth]{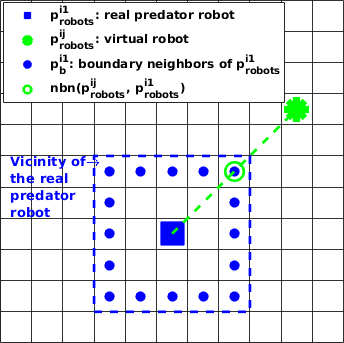}
  	\caption{Illustration of the function $nbn(p_{robots}^{ij}, p_{robots}^{i1})$ in Equation (\ref{eq_func_nbn}).}
  	\label{fig_func_nbn}		
\end{figure}

Two update rules are designed separately for virtual and real robots:

1. For a virtual robot $j$ ($j \in \{2,...,N_p\}$), the PSO update rules are as follows:

\begin{eqnarray}
v_{robots}^{ij} &=& nnd( w \cdot v_{robots}^{ij} + c_1 \cdot r_1 \cdot (pi_{robots}^{ij} - p_{robots}^{ij}) \nonumber \\
&& + c_2 \cdot r_2 \cdot (pg_{robots}^i - p_{robots}^{ij}) ) \label{eq_velocity_virtual}\\
p_{robots}^{ij} &=& nbn( (p_{robots}^{ij} + v_{robots}^{ij}), p_{robots}^{i1} ) \label{eq_position_virtual}
\end{eqnarray}
where 
\begin{equation}\label{eq_func_nnd_definite}
nnd(v) = \text{arg} \min\limits_{p_n \in SN} |\angle{p_n} - \angle{v}|
\end{equation}
and
\begin{equation}
SN = \{(1,0), (1,1), (0,1), (-1,1), (-1,0), (-1,-1), (0,-1), (1,-1) \}.
\end{equation}
$nnd(v)$ outputs one of the 8 unit vectors in $SN$ which has the minimum angle distance with the input velocity $v$. 
By using the function $nnd(\cdot)$, every robot can only move one step by one step. 
In this way, unlike the multi-steps case in a general PSO, the path planning and the worry about collisions in the half way to a destination are not ever necessary .
$v_{robots}^{ij}$ is the velocity for the $j$th individual (robot) in the $i$th subpopulation which has the position $p_{robots}^{ij}$. 
In addition, $pi_{robots}^{ij}$ is the individual historical best position for the $j$th individual (robot) in the $i$th subpopulation, 
while $pg_{robots}^i$ is the  global best position of the $i$th subpopulation.
The coefficient $w \in{R}$ is called the inertia weight, $c_1, c_2 \in{R^+}$, and $r_1, r_2$ are uniformly distributed random numbers in the range of $(0,1)$. 
Besides,

\begin{equation}\label{eq_func_nbn}
nbn(p_{robots}^{ij}, p_{robots}^{i1}) = 
\begin{cases}
\text{arg } \min\limits_{p_b^{i1}} |\angle{(p_b^{i1} - p_{robots}^{i1})} - \angle{(p_{robots}^{ij} - p_{robots}^{i1})}|, \\
\qquad\qquad\qquad \text{if }p_{robots}^{ij} \text{ is out of the vicinity of } p_{robots}^{i1} \\
p_{robots}^{ij},  \quad\qquad\text{otherwise}
\end{cases}
\end{equation}
is designed to output the nearest boundary neighbor $p_b^{i1}$ in the constrained vicinity of the real predator robot $p_{robots}^{i1}$. 
This function is illustrated in Figure \ref{fig_func_nbn}, where the constrained vicinity of $p_{robots}^{i1}$ is shown in a dashed square, which is determined as the minimum one that can accommodate the specified number of virtual robots.
Note that, the $nbn(p_{robots}^{ij}, p_{robots}^{i1})$ function in Equation (\ref{eq_position_virtual}) is very important because it can assure all the virtual robots $p_{robots}^{ij} (j \geq 2)$ are in the constrained vicinity of the real predator robot $p_{robots}^{i1}$, without which the subpopulation may lose the vicinity exploring capability for the real predator robot.

\begin{figure}[h!]
	\centering
	\includegraphics[width=\linewidth]{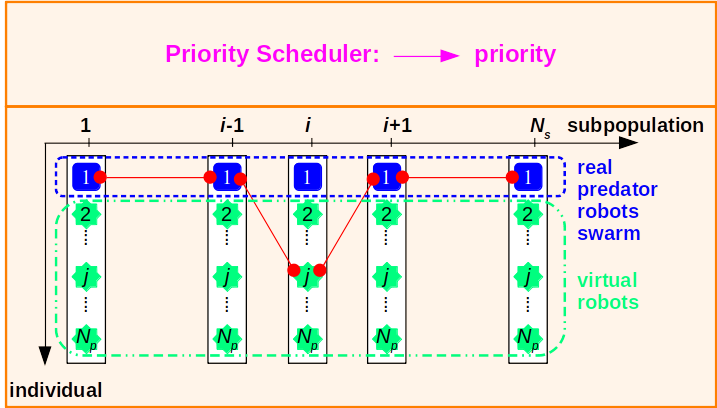}
	\caption{Illustration of the priority scheduler for the CC based fitness evaluation.}
	\label{fig_priority_schedular}
\end{figure} 

2. For the real robot ($j = 1$), the PSO update rules are as follows:
\begin{eqnarray}
v_{robots}^{i1} &=& nnd( pg_{robots}^i - p_{robots}^{i1} ) \label{eq_velocity_real}\\
p_{robots}^{i1} &=& p_{robots}^{i1} + v_{robots}^{i1} \label{eq_position_real}
\end{eqnarray}
So, the real predator robot does not need to perform any exploring task, but just quickly becomes the global best in its subpopulation. 

To summarize, by utilizing different optimization mechanisms for different kinds of robots, virtual robots are responsible for exploring and finding potential better positions in the vicinity of the real predator robot, while the real predator robot in each subpopulation just makes use of the achievements of the virtual robots and becomes the global best.

\subsubsection{Fitness evaluation}

From the practical point of view, no collisions of any two real robots are allowed. 
Since we have totally $N_s$ subpopulations in the cooperative coevolution population, 
a priority scheduler is used to coordinate among them, as shown in Figure \ref{fig_priority_schedular}. 
In particular, a priority scheduler will decide the order of movements among real robots, which can be the evolving order of subpopulations in asynchronous situations, and can also be used to coordinate, for example, two real robots when they both want to go to the same position in synchronous situations.

To be as simple as possible, here we let the priorities be in consistent with the indexes of the subpopulations. 
In other words, after the prey robot moves, the subpopulations evolve one-by-one and the newly updated real predator robot is counted into the dynamics of the environment for the fitness evaluations to the subsequent subpopulations.
So, if $k>h$, the predator $p^{k1}_{robots}$ always moves ahead of the predator $p^{h1}_{robots}$. 

To evaluate the fitness of the $j$th individual in the $i$th subpopulation, a complete solution should be first composed by replacing the $i$th real predator robot with $p^{ij}_{robots}$ from the real predator robots swarm:
\begin{equation*}
\left[p^{11}_{robots},...,p^{(i-1)1}_{robots},p^{ij}_{robots},p^{(i+1)1}_{robots},...,p^{N_s1}_{robots}\right].
\end{equation*}
Then, the fitness of a robot can be evaluated by equation (\ref{eq_fitness}), as shown in Figure \ref{fig_fitness_evaluation_ij}.

\subsubsection{Diversity maintenance mechanism}

When a swarm intelligence algorithm converges, all individuals may be attracted to the same position, no matter it is the global or local optimum. 
However, for the pursuit case here, the convergence of virtual robots in a subpopulation brings the disadvantage that the capability of exploring potentially better positions is getting worse.  
Therefore, if the number of unique virtual robots in a subpopulation is defined as the subpopulation diversity, the diversity of each subpopulation must be maintained to keep its exploring capability. 
Besides, due to the existence of unexpected deadlocks, suitable strategies should be integrated in the coordination algorithm to deal with  such problems. 

Based on the above ideas, we propose the diversity maintenance mechanisms which are performed as follows:
\begin{itemize}
\item Update the population in each generation based on the scheme that the fitness of the newly generated individual is not worse than its parental robot, which will guide the robot to explore more positions without harm to the fitness.  
\item Redistribute the virtual robots once the number of unique virtual robots positions in a subpopulation decreases below a threshold $T_v$, i.e., the subpopulation converges. 
That is, the subpopulation has found better solutions and all robots are attracted to the global best. 
In this situation, virtual robots should be redistributed to the space for better exploration. 
This strategy corresponds to the line 8-9 in Algorithm \ref{algorithm_CCPSO-R}.
\item Redistribute virtual robots once the real predator robot becomes the global best in the subpopulation. 
Because the role of virtual robots is to help the corresponding real predator robot to find better positions, once this real predator robot becomes the global best in its subpopulation, the object of virtual robots is reached and they should be redistributed to the space to find potential better positions for the real predator robot. 
This strategy corresponds to the line 12-13 in Algorithm \ref{algorithm_CCPSO-R}.
\item Add a random noise to the position of the real predator robot if it is not the global best in its subpopulation but abnormally keeps stills for a long time, in which it must have gotten stuck in a deadlock. 
This strategy corresponds to the line 14-16 in Algorithm \ref{algorithm_CCPSO-R}.
\item Add random noise to the positions of all the real predator robots if they converge when the prey robot has not been captured, the situation of which can be seen as that the swarm of predator robots gets trapped in a local optimum.
This strategy corresponds to the line 17-19 in Algorithm \ref{algorithm_CCPSO-R}.
\end{itemize}

\section{Experiments}\label{section_experiments}

In this section, two different experiments are presented.
Experiment 1 is conducted in a $30 \times 30$ grid world to verify the performance of the proposed CCPSO-R. 
Experiment 2 is to compare CCPSO-R with a representative dynamic path planning based pursuit algorithm MAPS \cite{Undeger2010}. 
From the experimental results , pros and cons of two different strategies can be seen in spite that CCPSO-R and MAPS are originally designed towards different capture definitions.

In particular, to verify the generality of algorithms, four types of preys are implemented. 
The prey robot initially locates in the center of the world, but behaves differently according to its type defined as follows:
\begin{itemize}
\item STILL PREY: the still prey keeps still in its initial position forever.
\item RANDOM PREY: the random prey randomly moves to a next position according to the uniform distribution.
\item LINEAR PREY: the linear prey initially chooses one of the 8 directions in which the number of predator robots is minimum, and moves in that direction in a straight line since then. 
Only when the prey locates on edges of the map, it will re-calculate a new direction according to the same criterion.
However, when the way of the linear prey is blocked by a predator, it cannot move any more but only wait for the other predator robots coming to encircle it.
\item SMARTER LINEAR PREY: the smarter linear prey, represented as linear\_smart, is very similar to the linear prey. The only difference is that when its way is blocked by a predator robot, it moves to an  unoccupied neighbor which has the minimum angle distance with its current direction and then it continues its movement in its previous direction if there are no obstacles.
\end{itemize}
From the above descriptions, the capabilities of the preys and the difficulties of encircling preys can be intuitively ranked as ``still prey $<$ random prey $<$ linear prey $<$ linear\_smart prey", which will be further verified by the following experiments. 

\subsection{Experiment 1}

\subsubsection{Experimental setup}
\label{section_experiment_1_setup}

To verify the scalability of CCPSO-R, various sizes of the swarm of predators, i.e., 4, 8, 12, 16 and 24, are used, from which we can expect the advantages originated from the swarm intelligence of the swarm of predator robots.  

The other implementation details are as follows: the initial real predator robots are deployed randomly in the whole grid world without overlapping;
the population size of each subpopulation is 20; 
the prey robot moves in 90\% of the time ensuring that predators move faster or a longer distance than the prey; in equation (\ref{eq_fit_repel}) $D_{min}=1$ which is the minimum secure distance between two robots; 
in equation (\ref{eq_position_virtual}) the parameters $w = 1, c_1 = c_2 = 2$ which are set as recommended in \cite{ShiPSO1998};
$T_v$ is 9 in the line 8 of Algorithm \ref{algorithm_CCPSO-R} which is the number of grids for a $3 \times 3$ vicinity; 
when the real robot is not the global best in its subpopulation but keeps still over 5 iterations we say that it gets trapped in a deadlock which corresponds to the line 14 of Algorithm \ref{algorithm_CCPSO-R}; 
when the swarm of predator robots keeps still over 10 iterations we say the swarm converged, and if the swarm has converged but the prey hasn't been captured we say that the swarm gets trapped in a local optimum which corresponds to the line 17 of Algorithm \ref{algorithm_CCPSO-R}.

In addition, for environmental changes such as real predator robot position change in other subpopulations, the current subpopulation needs to be re-evaluated as shown in the line 4 of Algorithm \ref{algorithm_CCPSO-R}, where the individual historical best position $pi^{ij}_{robots}$ will not be inherited, and the global best $pg^{i}_{robots}$ will be re-calculated.
This is because, although the experimental results of inheriting and not inheriting the individual historical memory $pi^{ij}_{robots}$ differ, it is hard to select either one due to their competitive performances.

As for the performance metrics, we use the number of successful captures, the average number of moves to capture the prey, and their standard deviations over 100 randomly generated test cases given the maximum 1000 time steps, the random seeds of which are set from 1 to 100.

\subsubsection{Simulation results}\label{section_simulation_results}

\begin{table}[h!]
	\begin{center}
	\caption{Number of captures, average number of moves, and their standard deviations to capture different preys with various number of predators out of 100 test cases.}	
	\label{tbl_experiment_statistic_results} 
	\begin{tabular}{|c|c|c|c|c|c|}\hline
    \multirow{2}{*}{No. of predators} & \multirow{2}{*}{Metrics}  & \multicolumn{4}{|c|}{Prey} \\
    \cline{3-6}
     & & Still & Random & Linear & Linear\_Smart \\
	\hline                               
    \multirow{3}{*}{4}  & No. of captures   & 100    & 100    & 100    & 100 \\
    \cline{2-6}
    	                & Avg. of moves & 30.450 & 49.840 &	46.900 & 204.060 \\
    \cline{2-6}    	               
        	            & Std. of moves & 19.943 & 36.867 &	31.886 & 198.927 \\    	               
	\hline\hline
    \multirow{3}{*}{8}  & No. of captures   & 100    &	100   & 100    & 100 \\
    \cline{2-6}
    	                & Avg. of moves & 22.220 & 33.780 &	42.240 & 121.820 \\
    \cline{2-6}    	               
        	            & Std. of moves & 13.384 & 23.039 & 46.583 & 111.922 \\  
	\hline\hline   
    \multirow{3}{*}{12} & No. of captures  & 100    & 100    & 100    & 100 \\
    \cline{2-6}
    	                & Avg. of moves & 20.470 & 24.520 &	30.190 & 76.780 \\
    \cline{2-6}    	               
        	            & Std. of moves & 11.364 & 13.414 &	24.943 & 72.839 \\  
	\hline\hline 	
    \multirow{3}{*}{16} & No. of captures & 100    & 100 & 100       & 100 \\
    \cline{2-6}
    	                & Avg. of moves & 17.360 & 18.360 & 25.620 & 49.850 \\
    \cline{2-6}    	               
        	            & Std. of moves & 9.648  & 11.277 & 23.560 & 50.048 \\  
	\hline\hline 	
    \multirow{3}{*}{24} & No. of captures & 100    & 100    & 100    & 100 \\
    \cline{2-6}
    	                & Avg. of moves & 15.060 & 14.060 &	19.670 & 35.400 \\
    \cline{2-6}    	               
        	            & Std. of moves & 10.688 & 6.151  &	21.879 & 32.588\\ 	     	           	
	\hline
	\end{tabular}
	\end{center}
\end{table}

\begin{figure}[h!]
	\centering
	\includegraphics[width=1.\linewidth]{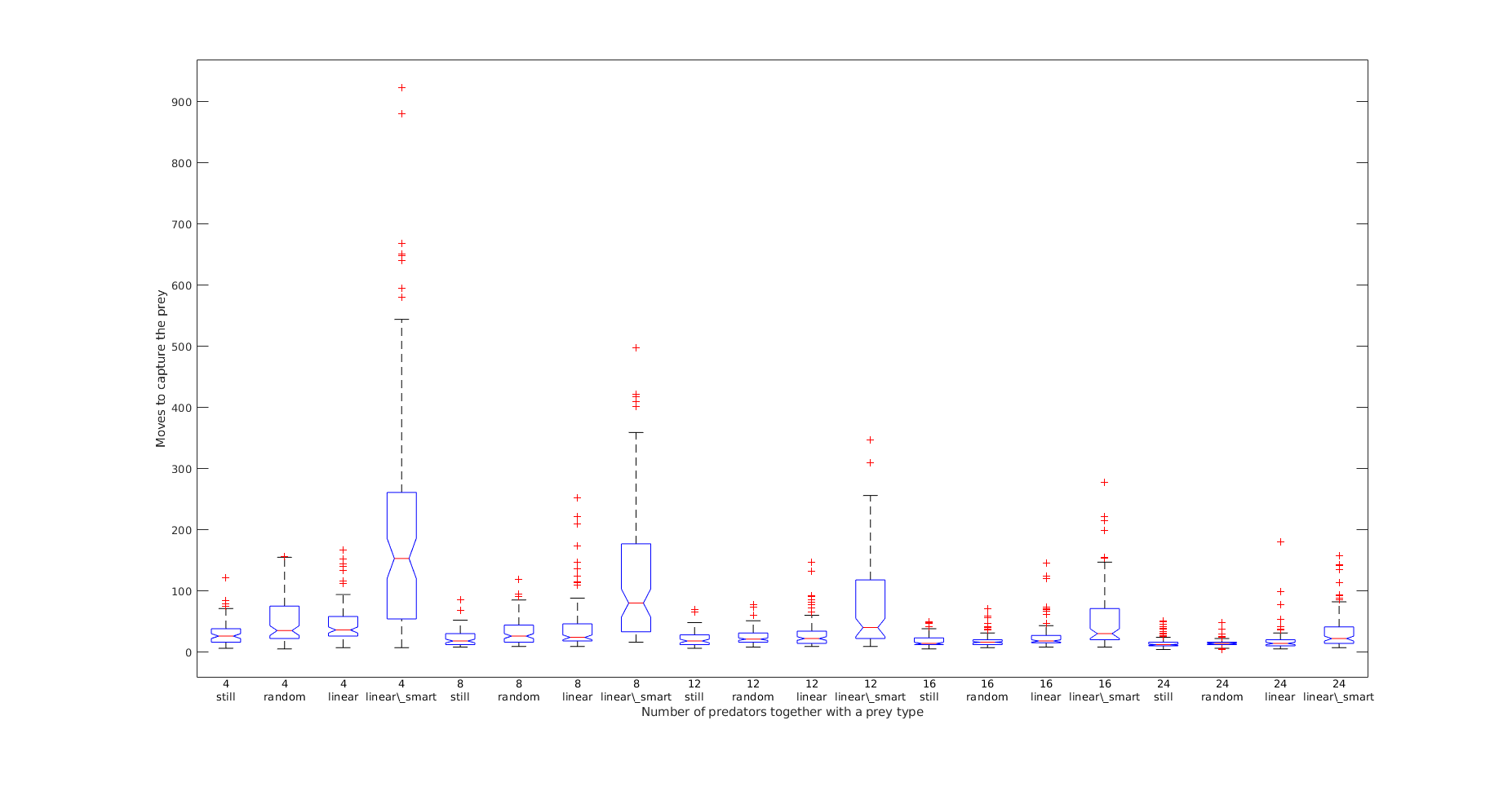}
	\caption{Box plot of the moves to capture a specific prey with a specific number of predator robots.}
	\label{fig_boxplot}
\end{figure}

\begin{figure}[h!]
	\centering
	\includegraphics[width=.6\linewidth]{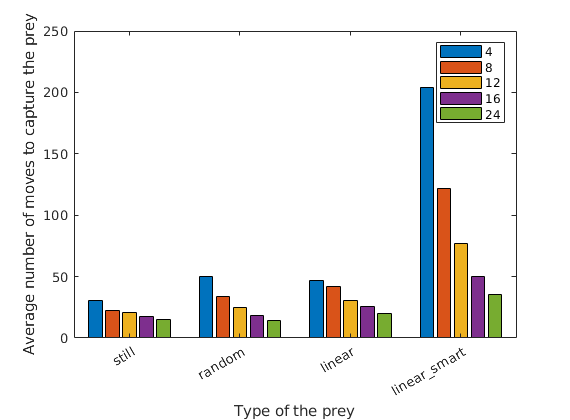}
	\caption{Bar graph of the average moves to capture a specific prey with a specific number of predator robots.}
	\label{fig_bar_group_prey}
\end{figure}

The simulation results are summarized in Table \ref{tbl_experiment_statistic_results},  from which it can be seen that CCPSO-R is reliable with the capture rate being 100\% in a limited time, no matter what type of the prey it is. 
As expected before, to a swarm of predator robots, the difficulties, in terms of the average number of moves, to capture each type of prey can be generally ranked as ``still prey $<$ random prey $<$ linear prey $<$ linear\_smart prey", which can be seen more clearly from Figure \ref{fig_boxplot}.
This conclusion is in consistent with the common opinion in literature  (such as \cite{HaynesEvolvingBehavioral} and \cite{Stone2000}) that compared with the random prey, the straight line moving prey is more effective because it breaks the movement locality.
Hence, the straight line moving prey is more difficult to be captured, which leads to the low capture rates in previous work, such as the manually designed methods \cite{korf1992simple,HaynesEvolvingBehavioral}, EA based method \cite{HaynesEvolvingBehavioral} and the case learning method \cite{HaynesLearning1998}. 

In addition, we show the data of Table \ref{tbl_experiment_statistic_results} in the manner of Figure \ref{fig_bar_group_prey}, from which an evident fact can be found that the more predator robots the more efficient the pursuit is. 
Besides, from the decreasing standard deviations as more real predator robots are involved, as shown in Figure \ref{fig_boxplot}, it can be concluded that with the swarm size of the predator robots gets larger, the pursuit performance is getting more and more stable and robust.

\begin{figure}[htbp]
	\centering
	\begin{subfigure}{.3\linewidth}
		\includegraphics[width=\linewidth]{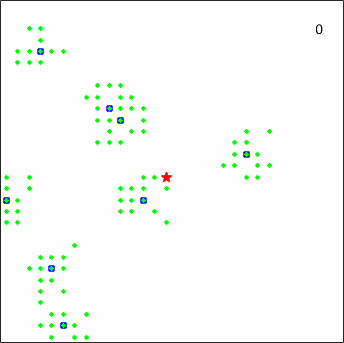}
  		\caption{}
	\end{subfigure} 
	\hfill
	\begin{subfigure}{.3\linewidth}
		\includegraphics[width=\linewidth]{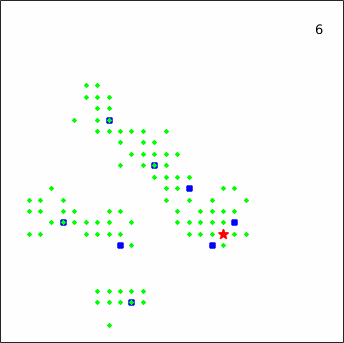}
  		\caption{}
	\end{subfigure}	
	\hfill
	\begin{subfigure}{.3\linewidth}
		\includegraphics[width=\linewidth]{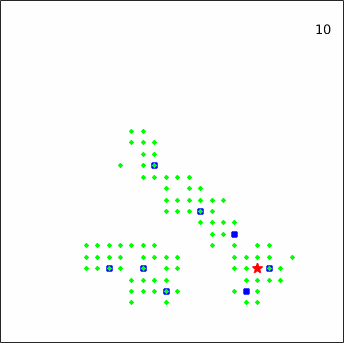}
  		\caption{}
	\end{subfigure}
	\\
	\begin{subfigure}{.3\linewidth}
		\includegraphics[width=\linewidth]{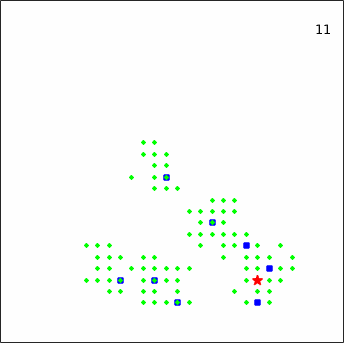}
  		\caption{}
	\end{subfigure} 
	\hfill
	\begin{subfigure}{.3\linewidth}
		\includegraphics[width=\linewidth]{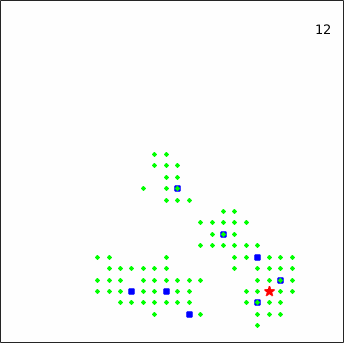}
  		\caption{}
	\end{subfigure}	
	\hfill
	\begin{subfigure}{.3\linewidth}
		\includegraphics[width=\linewidth]{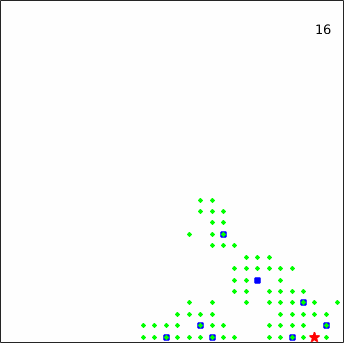}
  		\caption{}
	\end{subfigure}
	\\
	\begin{subfigure}{.3\linewidth}
		\includegraphics[width=\linewidth]{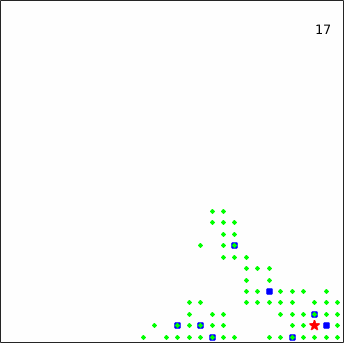}
  		\caption{}
	\end{subfigure} 
	\hfill
	\begin{subfigure}{.3\linewidth}
		\includegraphics[width=\linewidth]{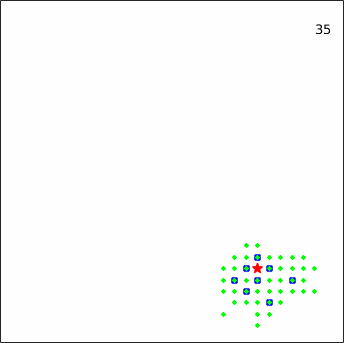}
  		\caption{}
	\end{subfigure}	
	\hfill
	\begin{subfigure}{.3\linewidth}
		\includegraphics[width=\linewidth]{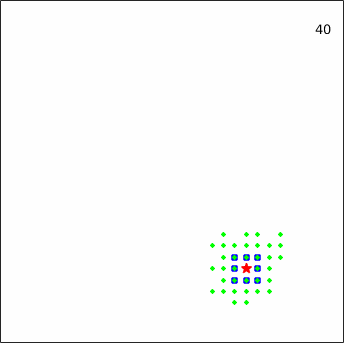}
  		\caption{}
	\end{subfigure}		
	
	\caption{Illustration of the pursuit process, taking the pursuit of a linear\_smart prey as an example. (a) is the initialization. From (a)-(b), the prey moves in the southeast direction in a straight line. In (c), the prey encounters an orthogonal real predator robot. So, in (d), the prey moves to a nearest unoccupied neighbor in the south. After that, from (d) to (e), the prey continues to move in its previous straight line direction. Until (f), the prey reaches an edge. So, in (g), the prey re-selects the north as its new escape direction. In (h), the prey is captured. And in (i), the predator robots swarm converge.}
	\label{fig_pursuit}
\end{figure}

To give a more intuitive impression of the pursuit process, several representative episodes taken from an experiment against the linear\_smart prey are displayed in Figure \ref{fig_pursuit}.

\subsubsection{Discussions}

The results achieved in Section \ref{section_simulation_results} can be explained from the algorithm's point of view.
First, as we treat the pursuit domain as an optimization problem, as long as the designed fitness function (or objective function) properly models the investigated problem, minimizing the fitness function will lead predators to a successful capture.
Second, on one hand, every step of a real predator robot is greedy since it moves to the best virtual robot position in its subpopulation; on the other hand, each step of a real predator is not totally greedy since the virtual robots exploration in its vicinity is not exhausted. 
So, predators may eventually capture the prey but the process may be slow.

Besides, in some of the past work, such as RL and  path planning approaches, the capture is defined differently as that the prey position is occupied by a predator, the further idealization and simplification of which result in an unpractical problem setup for robots applications.
This paper adopts another conventional definition that the prey is encircled by predators such that it cannot move any more, which considers both the collision avoidance and the safety of robots from the practical point of view.
However, to further validate the effectiveness of the proposed CCPSO-R, we will do the comparison in the next Sub-section.

\subsection{Experiment 2}\label{experiment_compare_with_maps}

For pursuit domain problems, path planning and task allocation based strategies are intuitive and may be the first solution that comes to one's mind.
Therefore, in this section, we compare the proposed CCPSO-R with a representative dynamic path planning based algorithm named MAPS  \cite{Undeger2010}.

\subsubsection{Experimental setup}

\begin{figure}[h!]
	\centering
	\begin{subfigure}{0.3\linewidth}
		\includegraphics[width=\linewidth]{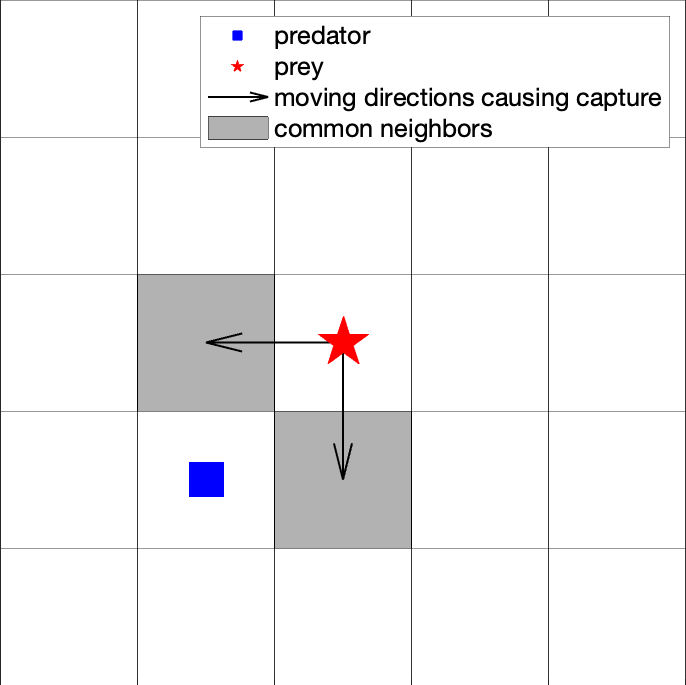}
		\caption{Example 1.}
  		\label{fig_capture_occupy_digonal}
	\end{subfigure}	
	\hfill
	\begin{subfigure}{0.3\linewidth}
		\includegraphics[width=\linewidth]{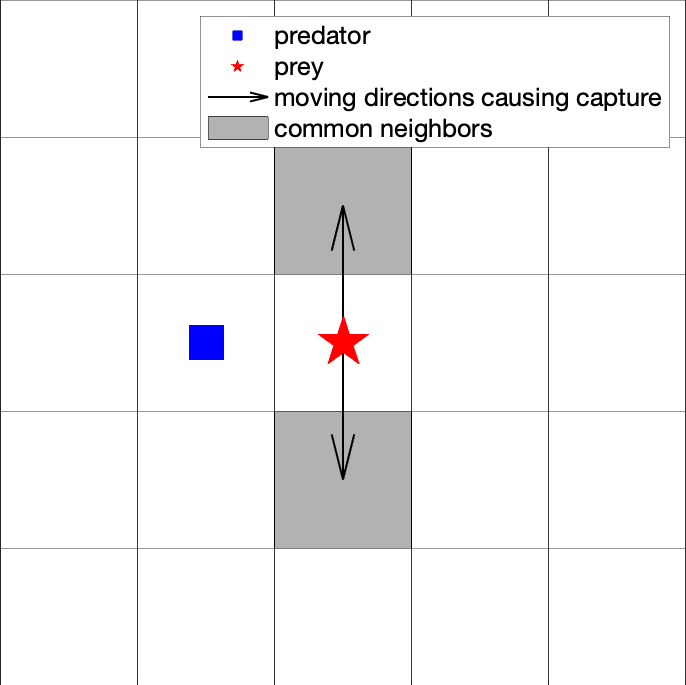}
		\caption{Example 2.}
  		\label{fig_capture_occupy_vertical}
	\end{subfigure}	
	\hfill
	\begin{subfigure}{0.3\linewidth}
		\includegraphics[width=\linewidth]{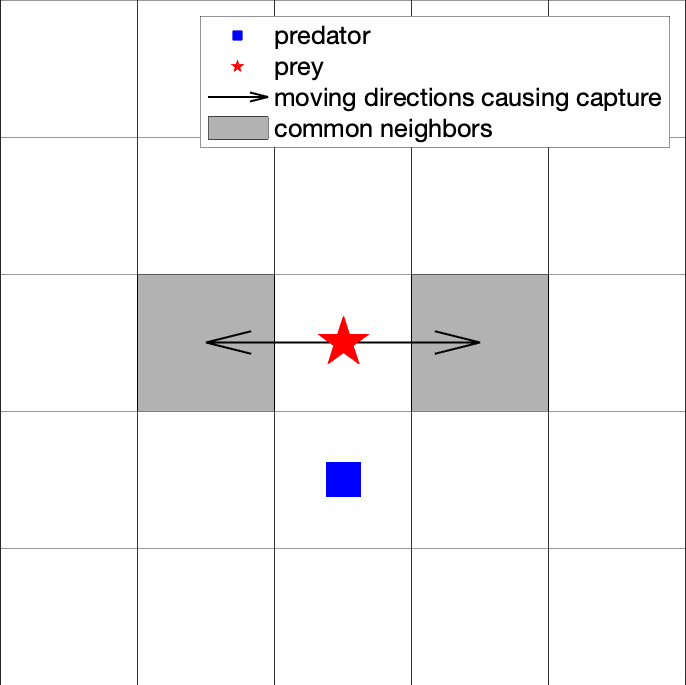}
		\caption{Example 3.}
  		\label{fig_capture_occupy_horizontal}
	\end{subfigure}		

	\caption{Illustration of the modified capture status of CCPSO-R used in the comparison with MAPS.}
	\label{fig_capture_occupy}
\end{figure}

All the experimental setups are the same as the Experiment 1 except the number of predator robots, the grid world sizes, and the way we calculate the capture status.
For fair comparisons, we modify the capture definition to the one adopted by MAPS that the position of the prey is occupied by any predator.
In particular, we know that in CCPSO-R, when predators cooperate to encircle the prey, there must be at least one such moment that the prey is adjacent to a predator and they get common neighbors as illustrated in Figure \ref{fig_capture_occupy}. 
Because diagonal obstacles are considered in our collision avoidance design, the number of the common neighbors is at most 2 except the prey's own position. 
If the prey keeps still or moves to anyone of the common neighbors the next moment, no matter it is compelled by other coordinated predators or just due to the simplicity of itself, the adjacent predator can definitely occupy the prey's new location since predators always move after the prey.
We can then say that the prey is captured at this moment.

\subsubsection{Simulation results}

\begin{table}[h!]
	\begin{center}
	\caption{Comparison results: Average number of moves and their standard deviations to capture different preys with various number of predators in different sizes of grid worlds out of 100 test cases.}	
	\label{tbl_experiment_compare_with_maps} 
	\begin{tabular}{|c|c|c||c|c||c|c||}
	\hline
    \multicolumn{3}{|c||}{Metrics} & \multicolumn{2}{|c||}{Avg. of moves} & \multicolumn{2}{|c||}{Std. of moves} \\   
	\hline
    \multicolumn{3}{|c||}{Number of predators}       & 4                         & 8                             & 4             & 8                \\
    \hline\hline
    \multicolumn{7}{|c|}{Size of the grid map: $30 \times 30$} \\
    \hline
    \multirow{8}{*}{Prey} &
    \multirow{2}{*}{Still} & MAPS         & 7.09 & 4.87 & 3.777 & 2.751  \\ 
    \cline{3-7}     
    {} & {}                          & CCPSO-R  & \textbf{6.5} & \textbf{4.66} & \textbf{2.819} & \textbf{2.271}  \\ 	            
	\cline{2-7}
   	{} & \multirow{2}{*}{Random} & MAPS & 9.01 & 6.78 & 6.973 & 5.868  \\ 
    \cline{3-7}     
    {} & {}                                 & CCPSO-R  & \textbf{8.3} & \textbf{6.04} & \textbf{3.335} & \textbf{2.828} \\ 	            
	\cline{2-7}
   	{} & \multirow{2}{*}{Linear} & MAPS     & 17.89 & 10.55 & \textbf{6.977} & 8.184  \\ 
    \cline{3-7}     
    {} & {}                                  & CCPSO-R  & \textbf{16.9} & \textbf{7.71} & 8.487 & \textbf{5.695} \\ 	            
	\cline{2-7}
   	{} & Linear                         & MAPS         & 17.84 & 10.55 & \textbf{7.102} & 8.211  \\ 
    \cline{3-7}     
    {} & Smart                         & CCPSO-R  & \textbf{16.92} & \textbf{7.75} & 8.474 & \textbf{5.723}  \\  
    \hline\hline
    \multicolumn{7}{|c|}{Size of the grid map: $150 \times 150$} \\
    \hline
    \multirow{8}{*}{Prey} &    
    \multirow{2}{*}{Still}         & MAPS         & 36.6 & 26.16 & 18.229 & 14.633  \\ 
    \cline{3-7}     
    {} & {}                                         & CCPSO-R  & \textbf{33.37} & \textbf{23.07} & \textbf{13.9} & \textbf{10.992} \\ 	            
	\cline{2-7}
   	{} & \multirow{2}{*}{Random} & MAPS         & 38.62 & 28.55 & 18.968 & 16.516 \\ 
    \cline{3-7}     
    {} & {}                                         & CCPSO-R & \textbf{35.84} & \textbf{26.07} & \textbf{13.654} & \textbf{11.401} \\ 	            
	\cline{2-7}
   	{} & \multirow{2}{*}{Linear}     & MAPS        & 99.33 & 64.29 & 44.576 & 37.918  \\ 
    \cline{3-7}     
    {} & {}                                         & CCPSO-R  & \textbf{82.75} & \textbf{41.32} & \textbf{30.274} & \textbf{29.684} \\ 	            
	\cline{2-7}
   	{} & Linear                                 & MAPS        & 99.19 & 64.37 & 44.849 & 37.814  \\ 
    \cline{3-7}     
    {} & Smart                         		   & CCPSO-R  & \textbf{82.84} & \textbf{41.44} & \textbf{30.192} &  \textbf{29.798} \\  
    \hline
	\end{tabular}
	\end{center}
\end{table}

\begin{table}[h!]
	\begin{center}
	\caption{Statistical test for CCPSO-R vs. MAPS using the t-test at the 5\% significance level.}	
	\label{tbl_experiment_significance_test} 
	\begin{tabular}{|c|c||c|c||c|c|}
	\hline
	\multicolumn{2}{|c||}{World size} & \multicolumn{2}{|c||}{$30 \times 30$} &  \multicolumn{2}{|c|}{$150 \times 150$} \\
	\hline
	No. of & \multirow{2}{*}{Prey} & \multirow{2}{*}{p-value} & \multirow{2}{*}{Significant} & \multirow{2}{*}{p-value} & \multirow{2}{*}{Significant}  \\
	predators & {} & {} & {} & {} & {} \\
	\hline
	\multirow{5}{*}{4} & Still & {0.013848} & {Yes} & {0.0062804} & {Yes} \\
	\cline{2-6} 
	{} 						  & Random & {0.13224} & {No} & {0.02533} & {Yes} \\
	\cline{2-6} 
	{} 						  & Linear & {0.10273} & {No} & {7.3641e-05} & {Yes} \\
	\cline{2-6} 
	\multirow{2}{*}{}  & Linear & \multirow{2}{*}{0.11762} & \multirow{2}{*}{No} & \multirow{2}{*}{9.8947e-05} & \multirow{2}{*}{Yes} \\
	{}                         & Smart & {} & {} & {} & {} \\
	\hline	
	\multirow{5}{*}{8} & Still & {0.13084} & {No} & {0.0017754} & {Yes} \\
	\cline{2-6} 
	{} 						  & Random & {0.11016} & {No} & {0.027238} & {Yes} \\
	\cline{2-6} 
	{} 						  & Linear & {0.00018443} & {Yes} & {1.3396e-11} & {Yes} \\
	\cline{2-6} 
	\multirow{2}{*}{}  & Linear & \multirow{2}{*}{0.00020334} & \multirow{2}{*}{Yes} & \multirow{2}{*}{1.1589e-11} & \multirow{2}{*}{Yes} \\		
	{}                         & Smart & {} & {} & {} & {} \\
	\hline
	\end{tabular}
	\end{center}
\end{table}

We run the two algorithms MAPS and CCPSO-R in the same randomly generated scenarios where the initial predators and prey positions are the same.
To see whether the map size influences the comparison results much, a bigger grid world size $150 \times 150$ together with a smaller grid map size $30 \times 30$ are applied with two predator swarm sizes 4 and 8.
Experimental results are summarized in Table \ref{tbl_experiment_compare_with_maps}.
Since both algorithms capture the prey $100\%$, we do not list the ``No. of captures" metric.
It is not hard to understand that generally with the increase of the predators number, less moves with smaller standard deviations are needed for a capture; and with the increase of the grid map size, values in these two metrics increase accordingly. 

To further validate the significance of the comparison results in Table \ref{tbl_experiment_compare_with_maps}, t-tests are conducted at the 5\% significance level, as shown in Table \ref{tbl_experiment_significance_test}.
It can be seen that CCPSO-R significantly outperforms MAPS in 11 out of the total 16 experimental scenarios where more predators are involved with smarter preys in bigger worlds.

\subsubsection{Discussions}

The comparison results show that 
although CCPSO-R is designed for the coordinated encirclement of the prey until every available orthogonal neighbor of the prey has been occupied by a predator, CCPSO-R also performs well in the position occupying based capture definition, which proves the effectiveness of the cooperative coevolutionary based coordination strategy.

In addition, compared with just occupying the same position of the prey with only one predator, occupying all the orthogonal neighbors of the prey simultaneously as uniformly as possible is more  complicated. 
Therefore, the proposed CCPSO-R can accomplish more complicated coordination tasks compared to MAPS.

On the other hand, as presented  in \cite{Undeger2010} that MAPS is a real-time  pursuit algorithm, its MATLAB version, which is rewritten by us from its original C++ codes, still runs faster than CCPSO-R. 
Because the calculation of the average number of moves per second over all the test cases has a high requirement on the runtime environment for fair comparison, we do not list this metric here.
But we can still get the following conclusions.
MAPS is faster.
However, one problem of it is that its performance is constrained by the number of predators.
Because one important step in MAPS is to assign the possible escape directions of the prey to every predator optimally by iterating every possible assignments, the combinatorial number of which is $(n - 1)!$ where $n$ is the number of predators.
With the increase of $n$, this combinatorial number will increase very fast and it is becoming less practical to get all the permutations at once, which also brings more burden to the memory.
This is also the reason that we only compare the simulation results with up to 8 predators.
But, for CCPSO-R, although it is not as efficient as MAPS, its scalability on the predator swarm size is much better.

Finally, despite the fact that both MAPS and CCPSO-R adopt the same sequential movement strategy that the prey always moves first and then the predators move one by one, another loop is embedded in the current CCPSO-R implementation that each real predator robot position can only be updated when all its corresponding virtual robots have been updated sequentially, as shown in line 5-10 of Algorithm \ref{algorithm_CCPSO-R}.
It is known that embedded loops are generally not expected for an efficient algorithm.
So, a parallel update and evaluation for the virtual robots should alleviate the efficiency constrain of CCPSO-R which will be investigated in future work.

\section{Conclusions}\label{section_conclusions}

This paper treated the pursuit domain as an optimization problem and presented the cooperative coevolutionary algorithm---CCPSO-R, which, for the first time, introduces the combination of the real robots and virtual robots into the correspondences between the individual representation of an EA and the robots in an application. 
Before the work in this paper, an individual in an EA will be assigned to a real robot. 
However, in the proposed CCPSO-R algorithm, only the first individual in each subpopulation corresponds to a real robot, while the rest individuals are all virtual robots, who act as a kind of action space for real robots by sampling and exploring their vicinities.

Besides, it should be noted that there are no fixed behavior rules for the swarm of predator robots. 
Instead, the swarm of robots is guided directly by the fitness function, which is designed in a modular manner by incorporating very limited domain knowledge. 
As one module, the collision avoidance consideration is integrated in the fitness function, which itself is another fitness function for repelling and can be versatile by tuning its parameter $D_{min}$. 
If the $D_{min}=1$, as it is in this paper, the robot swarm can capture the prey while moving without collisions. 

Finally, we tested the performance---the generality, stability and scalability of the proposed CCPSO-R with four types of preys---the still prey, the random prey, the linear prey, and the linear\_smart prey. 
Experimental results have been summarized based on 100 randomly generated test cases whose random seeds are set as 1-100 for their reproducibility. 
Based on these experiments, it can be concluded that the proposed CCPSO-R can always capture the prey stably and no additional modifications are needed under different scenarios.
In addition, a comparison with a representative dynamic path planning and task allocation based algorithm MAPS has also been conducted. Experimental results further prove the outstanding performance of the proposed CCPSO-R.

However, to be simple, the coordination priority scheduler was designed based on the subpopulation indexes, which indicates that the real predator robots move in a fixed sequential order. 
This may be unreasonable when it is better to firstly move one specific predator which blocks others' ways. 
In addition, predators move sequentially, rather than synchronously, will deteriorate the pursuit efficiency when the swarm of predators gets larger. 
Therefore, three works need to be done in future: one is to study the memory inheritance strategy in dynamic optimization problems as mentioned in Section \ref{section_experiment_1_setup}; one is to implement the parallel update and evaluation for the virtual robots; one is to improve the coordination scheduler towards the synchronous cooperation based on parallel computing by learning from experiences.

\section*{Acknowledgements} 
We would like to thank Dr. Cagatay Undeger for sharing their codes with us which made our comparison work feasible.
Additionally, Professor Chin-Teng Lin helped us a lot during this research work,
and Qiqi Duan gave kindly advice.
Finally, we want to give our deeply appreciation to all the anonymous reviewers for their constructive comments and suggestions.

\section*{References}

\bibliography{mybibfile}

\end{document}